\documentclass[letterpaper]{article} 
\usepackage{aaai2026}  
\usepackage{times}  
\usepackage{helvet}  
\usepackage{courier}  
\usepackage[hyphens]{url}  
\usepackage{graphicx} 
\usepackage{natbib}  
\usepackage{amsmath}
\usepackage{amssymb}
\usepackage{caption} 
\usepackage{comment}
\usepackage{algorithm}
\usepackage{algorithmic}

\usepackage{newfloat}
\usepackage{listings}
\DeclareCaptionStyle{ruled}{labelfont=normalfont,labelsep=colon,strut=off} 
\floatstyle{ruled}
\newfloat{listing}{tb}{lst}{}
\floatname{listing}{Listing}
\title{Measuring Stability Beyond Accuracy in Small Open-Source Medical Large Language Models for Pediatric Endocrinology}
\author{
    Vanessa D'Amario\textsuperscript{\rm 1}, 
    Randy Daniel\textsuperscript{\rm 1}, Alessandro Zanetti\textsuperscript{\rm 1}, Dhruv Edamadaka\textsuperscript{\rm 1}, Nitya Alaparthy\textsuperscript{\rm 1}, Joshua Tarkoff\textsuperscript{\rm 2} %
}
\affiliations{
    \textsuperscript{\rm 1}Nova Southeastern University\\ 3300 S University Dr, Fort Lauderdale, FL 33328 USA \\
    
    \textsuperscript{\rm 2}Nicklaus Children's Hospital\\ 3100 SW 62nd Ave, Miami, FL 33155 USA \\
    
    vdamario@nova.edu
}

\usepackage{bibentry}

\begin{document}

\maketitle

\begin{abstract}
Small open-source medical large language models (LLMs) offer promising opportunities for low-resource deployment and broader accessibility. However, their evaluation is often limited to accuracy on medical multiple choice question (MCQ) benchmarks, and lacks evaluation of consistency, robustness, or reasoning behavior. We use MCQ coupled to human evaluation and clinical review to assess six small open-source medical LLMs (HuatuoGPT-o1~\cite{chen2024huatuogpt}, Diabetica-7B~\cite{wei2024adapted}, Diabetica-o1~\cite{wei2024adapted}, Meditron3-8B~\cite{sallinen2025llama}, MedFound-7B~\cite{liu2025generalist}, and ClinicaGPT-base-zh~\cite{wang2023clinicalgpt}) in pediatric endocrinology. In deterministic settings, we examine the effect of prompt variation on models' output and self-assessment bias. In stochastic settings, we evaluate output variability and investigate the relationship between consistency and correctness. HuatuoGPT-o1-8B achieved the highest performance. The results show that high consistency across the model response is not an indicator of correctness, although HuatuoGPT-o1-8B showed the highest consistency rate. When tasked with selecting correct reasoning, both HuatuoGPT-o1-8B and Diabetica-o1 exhibit self-assessment bias and dependency on the order of the candidate explanations. Expert review of incorrect reasoning rationales identified a mix of clinically acceptable responses and clinical oversight. 
We further show that system-level perturbations, such as differences in CUDA builds, can yield statistically significant shifts in model output despite stable accuracy. This work demonstrates that small, semantically negligible prompt perturbations lead to divergent outputs, raising concerns about reproducibility of LLM-based evaluations and highlights the output variability under different stochastic regimes, emphasizing the need of a broader diagnostic framework to understand potential pitfalls in real-world clinical decision support scenarios.
\end{abstract}


\section{Introduction}

Large Language Models (LLMs) hold significant potential in healthcare, particularly to improve chronic disease management and clinical decision support \cite{tu2024towards, mcduff2025towards}. High-performance medical LLMs have showed strong results on different aspects of clinical care, from diagnostics to patient interfacing~\cite{singhal2025toward, liu2025generalist, mukherjee2024polaris}. However, these models are often large, exceeding $70$ billions parameters, and proprietary, making them inaccessible for independent validation or adaptation. They become impracticable solutions in resource-constrained or low-income healthcare settings.  

In contrast, small open-source medical LLMs offer the potential for lightweight, locally deployable, and transparent tools that could reduce barriers to access. Nonetheless, their reliability and reproducibility have not been systematically addressed. A critical question to address is understanding if these models, as released and without fine-tuning, encode sufficient clinical knowledge to support real-world use. 

Pediatric endocrinology presents a unique testbed for the evaluation of medical LLMs. Pediatric endocrinology is facing growing patient volume, longer waiting times, and a larger range of conditions in the last years \cite{kumar2021pediatric, 10.1542/peds.2023-063678J}.  
At the same time, access to pediatric endocrinology expertise is limited in many regions \cite{10.1542/peds.2023-063678J}, creating a potential role for AI-driven decision support, provided that they can deliver trustworthy outputs. 

In this study, we explore this question by evaluating a selection of small (under 10 billion parameters) previously fine-tuned open-source medical LLMs in pediatric endocrinology. Given the lack of evaluation of medical LLMs in this field, our work fills this gap by systematically assessing model behavior on a curated set of pediatric endocrinology cases, offering insights into both the capabilities and limitations of LLM tools in this clinical subspecialty.

To evaluate model performance in pediatric endocrinology, we leverage Multiple Choice Questions (MCQs) drawn from real-world clinical content. MCQs are a widely used strategy for evaluating medical LLMs, particularly small, domain-specific models. While convenient and well-structured for automated scoring and scalable benchmarking, MCQ-based evaluation has well-documented limitations. Question difficulty varies widely, distractors may lack clinical plausibility, and accuracy alone often obscures reasoning quality, robustness, or consistency. Recent studies outside of the healthcare context~\cite{li2024can, khatun2024study} have also highlighted response instability among LLMs in MCQ settings and self-assessment bias~\cite{xu2024pride, wataoka2024self, panickssery2024llm}.

Despite these challenges, MCQs remain one of the most accessible formats for systematic model evaluation. In this work, we embrace their practicality while addressing their shortcomings: we propose a multidimensional evaluation framework that goes beyond standard accuracy-based evaluation and measures how stable model outputs are under clinically irrelevant input variations, parameter changes, and computational variability. Specifically, we evaluate six small medical LLMs and assess \emph{consistency}, by comparing decisions across prompt variations, including syntactical variations and the removal of answer option labels. We also evaluate \emph{response consistency and correctness} across stochastic sampling, where the degree of stochasticity of the inference process is determined by different temperature values and \emph{self-assessment bias}, by presenting models with a choice between their own rationale and a gold-standard clinical explanation. We lastly investigate \emph{numerical stability}, examining how changes in software environments affect model outputs despite identical inputs. 

This multidimensional analysis provides a structured diagnostic of small medical LLM readiness and exposes failure modes that accuracy metrics alone would not be able to capture. Our goal is not only to compare models but to define reproducibility-oriented evaluation criteria for trustworthy, accessible LLMs in medicine. In doing so, we highlight which models show the greatest promise for practical clinical integration and identify areas requiring further improvement.

\section{Materials and Methods}
\subsection{Multiple Choice Questions from Pediatric ESAP 2021-2022}
To evaluate the reasoning capabilities of small open-source medical LLMs we rely on the Pediatric Endocrine Self-Assessment Program 2021-2022 (ESAP) \cite{pesce2021pediatric}. The program is a question bank containing 100 among clinical cases and factual questions based on clinical guidelines. Clinical cases items present a patient scenario, including relevant medical history, examination findings, and laboratory results, and typically ask to determine the most likely diagnosis or the best course of action in patient management. Other more factual items focus on guideline-based knowledge such as indications for specific medications, appropriate monitoring strategies, or recommended referral to a subspecialist. Each ESAP item includes a description of the clinical case, a question, five possible answers (each comes with a letter from A to E), an educational objective, the correct answer, and the clinical explanation for each answer. Drawing on the expertise of leading pediatric endocrinologists, the program is highly technical and aimed at medical specialists. The program's learning objectives include recognizing clinical features of endocrine, growth, and metabolic disorders; applying current diagnostic and treatment options; identifying risk factors; and evaluating endocrine manifestations of systemic diseases. As such, we considered it as an optimal starting point for knowledge assessment in pediatric endocrinology.

We excluded nine ESAP items from the analysis, since those rely on images and graphs. 
For each clinical case described, we extracted the following fields: \texttt{<description>}, \texttt{<question>}, \texttt{<options>}, \texttt{<correct answer>}, and \texttt{<explanation>}. 


\subsection{Small Open-Source Medical LLMs}
We evaluate six small-sized, open-source medical LLMs designed for biomedical reasoning. 

\emph{HuatuoGPT-o1}~\cite{chen2024huatuogpt} is based on the LLaMA 3.1 8B architecture \cite{dubey2024llama} and is trained using continued pre-training and supervised fine-tuning in a bilingual corpus of 1.1TB (Chinese and English), with benchmarks including MedQA, MedMCQA, CMB, CMExam, and CMMLU. 

\emph{Diabetica-7B}~\cite{wei2024adapted} uses the Qwen2-7B-Instruct model \cite{yang2024qwen2technicalreport}, fine-tuned through low-rank adaptation and self-distillation on public medical QA sets, SFT datasets, and internal diabetes-specific data, evaluated through MCQs, fill-in-the-blank and dialogue tasks. 

\emph{Diabetica-o1} is the result of self-distillation of Diabetica-7B. 

\emph{Meditron3-8B}~\cite{sallinen2025llama}, using LLaMA 3.1-8B, underwent a continuous pre-training and supervised instruction fine-tuning in PubMed Central, validated medical textbooks, and more than 46,000 clinical practice guidelines, and was tested on MedQA, MedMCQA, and PubMedQA. 

\emph{MedFound-7B}~\cite{liu2025generalist} and \emph{ClinicalGPT-based-zh}~\cite{wang2023clinicalgpt} are both based on the BLOOM-7B~\cite{workshop2022bloom}. The latter was developed through supervised fine-tuning and reinforcement learning using a range of Chinese electronic health records, multi-turn patient-doctor dialogues, standardized medical examination questions, and question-answer pairs derived from structured medical knowledge graphs. 

\subsection{Experimental Design}

To assess the stability and reasoning capabilities of small open-source medical LLMs, we prepared our dataset from the Pediatric ESAP 2021-2022 as a stepping-stone in the design of three main experiments.
We describe the inferential setup for all the experiments in Appendix \emph{Experimental Setup/Inference and Hyperparameters}.


\subsubsection{Experiment 1: LLMs Accuracy and Response Stability to Prompt Variation}
We evaluate the performance of the models under deterministic conditions (temperature $T=0$). For each ESAP item, including the clinical description, question and multiple-choice options, each model is instructed to select the correct option and provide supporting clinical reasoning. Three prompt strategies are tested: (1) prompt A (Appendix \emph{Prompting/Prompts in Experiment 1}), (2) prompt B, which introduces minor syntactical variations to prompt A and is used to assess prompt sensitivity; and (3) prompt A without letter token, which is identical to prompt A except that the option letter (A to E) are omitted from the \texttt{<options>} field, separating each option with a semicolon. 

Beyond accuracy, the main objective of this experiment is to assess the stability of model outputs to prompt variations. We apply McNemar's exact test to compare accuracy across (a) syntactic variations (prompt A vs prompt B) and (b) letter omission (prompt A vs prompt A without letters). To capture variability in model behavior independently of correctness, we also compute the pairwise response match-rate and Cohen's $\kappa$ coefficient, each accompanied by $95\%$ confidence intervals obtained via Wilson and bootstrap methods ($B=10000$), respectively. A Stuart-Maxwell test is performed to evaluate whether the distribution of categorical responses differs significantly between prompt strategies, providing complementary evidence of systematic response shift beyond overall accuracy.

\subsubsection{Experiment 2: Stability in non-deterministic setting}
We examine the reliability of the prediction under stochastic decoding, for increasing temperature values. Despite finding stable performance of commercial models (ChatGPT, Llama 70B) across different temperatures~\cite{renze2024effect, patel2024exploring}, recent studies have showed how model performance degrades for LLMs of limited size ($<13$ B) over $T\ge 1$, especially for tasks that require instruction following~\cite{li2025exploring}. In our experiments, we evaluate the output to prompt A in three different settings $T=[0.3, 0.6, 1.0]$. We focus on the models that perform the best in Experiment 1. We define \emph{consistency} as the frequency of \emph{majority vote} and \emph{correctness} as the frequency of correct answers in ten runs. This setup allows us to assess whether high-confidence predictions consistently correspond to correct answers. We also determine the model stability and best performing model at different stochastic regimes.

\subsubsection{Experiment 3: Self-Bias and Discerning Gold-Standard Reasoning}
We provide the LLM with the ESAP item description, question, and multiple choice options. We ask the LLM to select the most clinically correct solution between two candidate explanations, with one being the model’s own output from Experiment 1, the other being the gold-standard explanation authored by ESAP instructors (prompt in App. \emph{Prompting/Prompt in Experiment 3}). For this task, we used both responses from prompt A and prompt B. We additionally ask the LLM to output a degree of agreement between the two explanations on a scale from 0 to 5. To evaluate the presence of positional bias, we prompted the model in two ways: once with the gold-standard explanation listed first and the model’s reasoning second, and then with reversed order. 
During the analysis, we focus on the ESAP items where the model selected an incorrect response in Experiment 1, since those necessarily diverge from the indicated best answer. Incorrect might reflect factually wrong statements, as well as overly cautious management plan, which do not align with clinical expectations.

Given the time-consuming nature of this task, the clinical review focuses on the most critical cases. The pediatric endocrinology expert is asked to evaluate the clinical reasoning of the best-performing model on items that were consistently answered incorrectly, independent of the position bias. The evaluation is based on one of the items of the CLEVER framework proposed for clinical reasoning, in ~\cite{liu2025generalist}, using a Likert scale from 1 to 5. Clinical reasoning is defined as the alignment of the LLM's content with the diagnostic reasoning process used in medical practice.

\subsubsection{Reproducibility Considerations in Model Implementation}
Experiment 1-2-3 are run on a single NVIDIA L40S (CUDA 11.8, cuBLAS 11.11.3.6, cuDNN 9.1.0.70, NCCL 2.21.5). To assess implementation-level reproducibility, the best performing models in Experiment 1 were re-run on prompt A on a second system NVIDIA RTX 5090 (CUDA 12.8, cuBLAS 12.8.3.14, cuDNN 9.7.1.26, NCCL 2.26.2). Both systems used the Hugging Face \texttt{transformers (4.51.3)} library under deterministic conditions (\texttt{do\_sample=False}), with identical software dependencies and decoding parameters.  Although both setups were functionally equivalent, it is common knowledge that GPU-level floating-point operations and cuBLAS/cuDNN kernel differences across CUDA versions can introduce minor nondeterminism in matrix computations~\cite{he2025nondeterminism}. We include this cross-system comparison to quantify the impact of hardware-software stack variability on model stability.

\subsubsection{Output Evaluation}
In our initial design, we planned to automate the output evaluation using GPT-4o, but we abandoned this approach due to the lack of consistency in GPT-4o responses. The output evaluation is conducted separately by three non-medical reviewers. 
Beyond the five standard ESAP options (A-E), we include two further categories: \emph{Multiple Selection} and \emph{Hallucinate or None}. We resume to \emph{Multiple Selection} whenever a model selects two or more options. In cases where the model response includes conflicting, missing, or ambiguous selections (\emph{e.g.,} indicating one answer, then contradicting itself), the output is marked as \emph{Hallucinate or None}. \emph{Hallucinate or None} reflects the difficulty of non-endocrinology reviewers to draw a line between valid but non-assertive medical verbiage and subtle medical errors.

For instances where non-medical reviewers could not attribute one of the categories, two clinical experts, one in emergency medicine and one in pediatric endocrinology, are consulted to assist with classification. 

The code to reproduce LLMs responses and results is available at \\ \url{https://github.com/vanessadamario/SmallMedLLMs-PedEndo}

\section{Results}

\begin{table*}[!h]
\centering  
\begin{tabular}{lcc|cc|cc} 
\hline
\textbf{Model} & \multicolumn{2}{c}{\textbf{Prompt A}} &
\multicolumn{2}{c}{\textbf{Prompt B (similar to A)}} &
\multicolumn{2}{c}{\textbf{Prompt A without Letter Token}} \\
 & accuracy (95\% CI) & \%u & accuracy (95\% CI) & \%u & accuracy (95\% CI) & \%u\\
\hline
\hline
\textbf{HuatuoGPT-o1-8B}     & $\mathbf{0.35[0.26,0.45]}$ & $\mathbf{100}$ & $\mathbf{0.35[0.26,0.45]}$ & $\mathbf{100}$ & $\mathbf{0.33[0.24,0.43]}$ & $\mathbf{96.7}$ \\
Diabetica-o1                     & $0.33[0.24,0.43]$          & $100$          & $0.34[0.25,0.44]$          & $100$          & $0.27[0.19,0.37]$          & $94.5$          \\
Diabetica-7B                    & $0.30[0.21,0.40]$          & $98.9$         & $0.32[0.23,0.42]$          & $100$          & $0.27[0.19,0.37]$          & $87.9$          \\
Meditron3-8B                  & $0.33[0.24,0.43]$          & $97.8$         & $-$           & $-$            & $0.34[0.25,0.44]$          & $90.1$          \\
MedFound-7B               & $0.04[0.02,0.11]$        & $18.6$         & $0.12[0.07,0.20]$          & $34.1$         & $0.04[0.02,0.11]$           & $22.0$          \\
ClinicalGPT-base-zh        & $0.20[0.13,0.29]$          & $79.1$         & $0.20[0.13,0.29]$          & $94.5$         & $0.20[0.13,0.29]$         & $79.1$          \\ 
\hline 
\end{tabular}
\caption{Assessment of correctness and usability in experiment 1. \textit{Legend.} For the three prompting strategies we report \textbf{accuracy} and 95\% confidence interval. \textbf{\%u}: percentage of usable responses, where the decision falls in one of the five ESAP options. Left section: prompt A. Center section: prompt B, similar to prompt A. Right section: prompt A without letter (A-E) in the multiple options.}
\label{tab:experiment1_accuracy}
\end{table*}

\begin{table*}[!h]
\centering  
\begin{tabular}{l|ccc|ccc} 
\hline
\textbf{Model} & \multicolumn{3}{c}{\textbf{Prompt A vs Prompt B}} &
\multicolumn{3}{c}{\textbf{Prompt A with vs without Letter Token}} \\
& Stuart- &$\kappa$   & match-rate & Stuart- & $\kappa$ & match-rate  \\
& Maxwell & 95\% CI    & 95\% CI     & Maxwell &  95\% CI  & 95\% CI      \\
\hline
\hline
\textbf{HuatuoGPT-o1-8B}  & $<10^{-4}$ & $0.55$         & $0.64$           & $<10^{-4}$ & $0.35$           & $0.48$   \\ 
                      &            & $[0.42, 0.67]$ & $[0.53,0.73]$    &            & $[0.23,0.48]$    & $[0.38,0.58]$   \\ 
\hline
Diabetica-o1              & $<10^{-4}$    & $0.40$        & $0.52$         &$<10^{-4}$   & $0.35$           & $0.48$   \\  
                          &               & $[0.27,0.52]$ & $[0.42,0.62]$  &           & $[0.23,0.48]$    & $[0.38,0.58]$   \\      
\hline
Diabetica-7B              &$<10^{-4}$  & $0.63$        & $0.70$         &$<10^{-4}$   & $0.38$           & $0.51$   \\
                          &            & $[0.51,0.74]$ & $[0.60,0.79]$  &             & $[0.26,0.51]$    & $[0.40,0.61]$   \\
\hline
Meditron3-8B              &$-$                  &$-$                 &$-$            & $0.006$      & $0.19$           & $0.35$   \\
                          &                   &                   &             &             & $[0.07,0.31]$    & $[0.26,0.45]$   \\
\hline
MedFound-7B               &$-$      &$-0.18$         &$0.05$        & $0.72$  &$-0.20$          & $0.04$   \\
                          &            &$[-0.24,-0.11]$ &$[0.02,0.12]$ &             &$[-0.24,-0.14]$  & $[0.02,0.11]$   \\
\hline
ClinicalGPT-base-zh       &$0.19$       & $0.15$        &$0.32$          &$0.005$      & $0.18$           &$0.34$    \\
                          &            & $[0.04,0.27]$ &$[0.23,0.42]$   &             & $[0.05,0.30]$    &$[0.25,0.44]$    \\
\hline
\end{tabular}
\caption{Assessment of stability and reproducibility in experiment 1. McNemar's test results in p-value $>0.4$ across all comparisons with exception of MedFound-7B Prompt A vs Prompt B. No significant differences between accuracies across pairwise comparison of prompting strategy. Beyond accuracy, stability is measured by Stuart-Maxwell test, to compare model output across different prompting strategies, under the hypothesis of no difference. p-value show significant differences. Cohen's $\kappa$ coefficient and match-rate are reported with their respective confidence intervals.}
\label{tab:experiment1_stability}
\end{table*}
Before showing the results of each experiment, we comment on the overall performance and hallucinations observed during output evaluation.

Among the models, \textit{HuatuoGPT-o1-8B}, \textit{Diabetica-7B}, and \textit{Meditron3-8B} exhibited the most consistent reasoning patterns. Their responses generally adhered to the multiple-choice format, often discussing the given options and weighing their clinical likelihood before reaching a conclusion. HuatuoGPT-o1-8B occasionally reversed this order, presenting a final answer first, followed by a discussion of each option as requested in the prompt. In contrast, \textit{Diabetica-o1} produced outputs that were more difficult to interpret. While the initial portion of the response is often coherent, the output frequently degenerates into simulated dialogues between a fictitious user and an assistant. These interactions included follow-up questions and, in some cases, newly generated MCQs. Hallucinations also included emojis, Chinese ideograms, LaTeX or Python code fragments, or statistical references disconnected from the task at hand. For the evaluation, we considered the first part of the output, and discarded what was unrelated to the analysis of the clinical scenario. \textit{MedFound-7B}'s output also showed signs of degeneration. The model frequently hallucinated content; when not hallucinating, it was hesitant to provide a definitive answer and tended to suggest a wide array of additional diagnostic tests before committing to a conclusion. The \textit{ClinicalGPT-base-zh} model produced shorter responses, typically only a few sentences, and rarely addressed all options. It often displayed a strong preference for a single option without discussing alternatives. For quantitative evaluations of option distributions, see App. \emph{Output Evaluation}.
 

\subsection{LLMs Performance, Prompt Stability, and Agreement}
In Table~\ref{tab:experiment1_accuracy}, for each LLM we report accuracy, confidence interval, and usability rate. The left, middle, and right panels show respectively the performance under prompt A, prompt B, and prompt A without letter token. 


\begin{figure*}[!ht]
    \centering
    \includegraphics[width=1.0\linewidth]{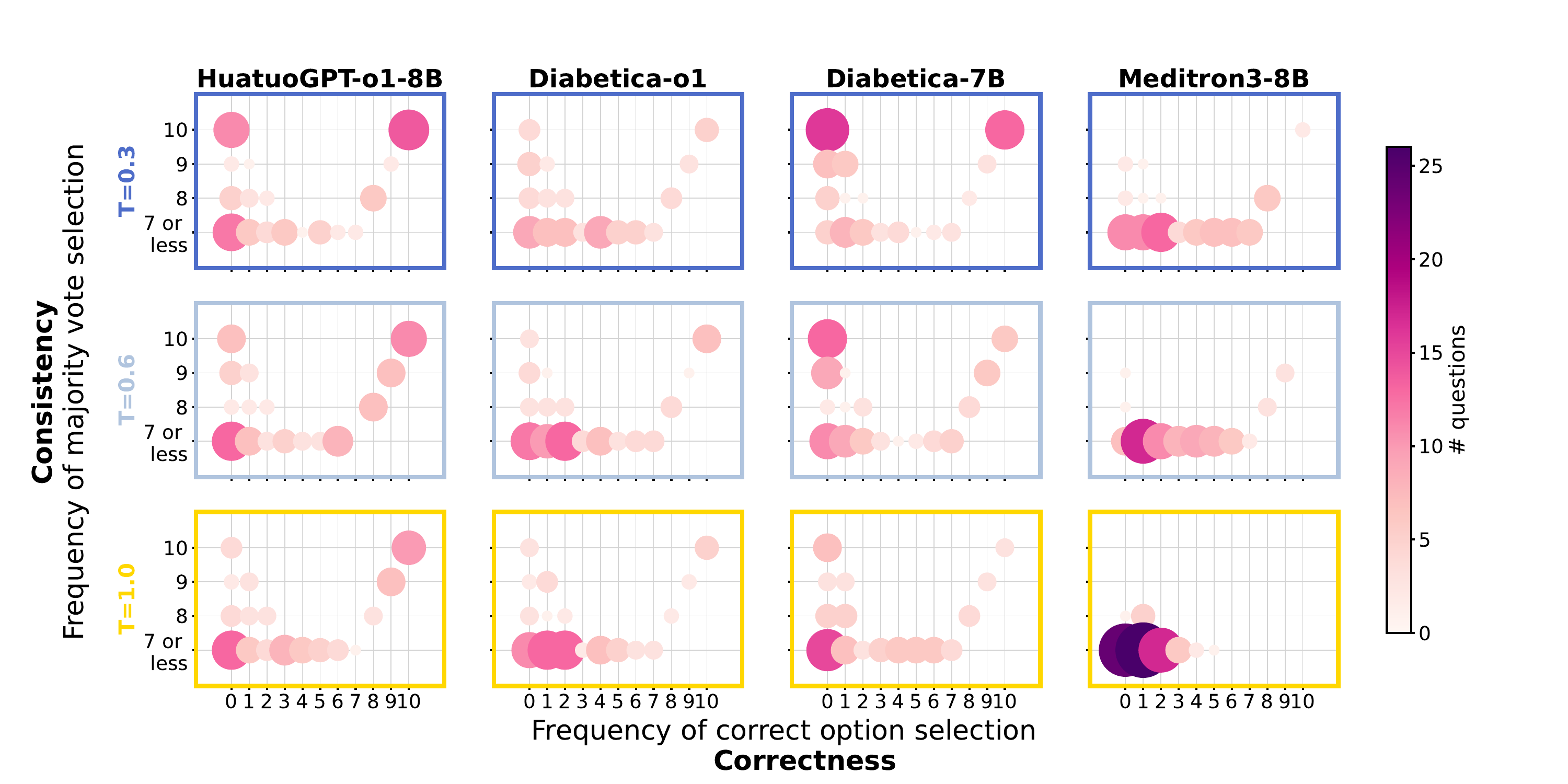}
    \caption{Consistency and correctness in the stochastic setting. From left to right, representations for HuatuoGPT-o1-8B, Diabetica-o1, Diabetica-7B, and Meditron 3-8B. From top to bottom, results across different temperature setting with $T=0.3$ (blue frame), $T=0.6$ (gray frame), and $T=1.0$ (yellow frame). In each plot, the $y$-axis reports consistency as majority vote selected $\ge 7$ times, up to ten. Correctness varies from 0 to total number of runs ($=10)$, on the $x$-axis. The size and color intensity of the blob show the number of cases that fall into a specific category. Ideally, a perfect-scoring model would show all the density concentrated in a single blob at the top right corner of the plot. We excluded all ESAP items where the model lacks a majority class. }
    \label{fig:Figure1a}
\end{figure*}
\begin{figure*} 
    \centering
    \includegraphics[width=.8\linewidth]{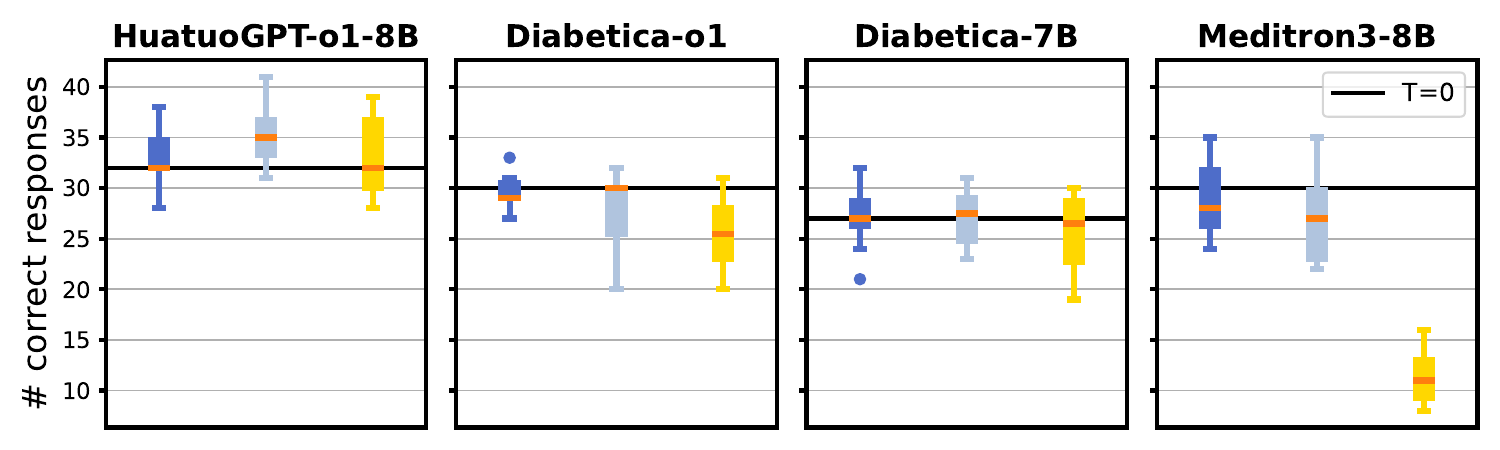}
    \caption{Box-plot showing the distribution of correct responses across runs, at three temperatures (color codes as in Figure \ref{fig:Figure1a}). The solid horizontal line shows performance under deterministic condition for each model. }
    \label{fig:Figure1b}
\end{figure*}

\emph{HuatuoGPT-o1-8B} emerges as the model with the highest number of correct responses across prompt A and prompt B. 
Although HuatuoGPT-o1-8B is slightly more accurate than Diabetica-o1, a closer look at the LLMs outputs reveals important differences. A breakdown on Diabetica-o1 performance shows that under prompt A and B the model outputs respectively only 12 and 13 fully interpretable answers, the output contains emojis in 12 and 14 cases, the output is truncated in 34 and 45 cases, the output contains LaTeX, Python, or Markdown code in 27 and 23 cases, and Chinese characters in 3 cases, across the two prompts.
As expected, Diabetica-7B performs slightly worse than Diabetica-o1. 
Meditron3-8B shows relatively high performance compared to other models for prompt A, but it fails to produce an output under prompt B. 
Models with lower overall performance, such as ClinicalGPT-base-zh, exhibit close to chance performance ($\text{chance}=18.2$ correct answer) and a strong bias toward a particular response (often option A). For further discussion, refer to App. \emph{Output Evaluation}.

Evaluations of model stability are in Table \ref{tab:experiment1_stability}. The table does not contain statistics and p-value for the McNemar's test. This resulted in p-values $>0.4$ across all models and configurations (prompt A vs prompt B, prompt A with letter option and without), indicating that overall accuracy remained stable despite syntactic and formatting changes. Nevertheless, subsequent analyses (Stuart-Maxwell test, Cohen's $\kappa$ coefficient, and match-rate) revealed variations in response patterns, suggesting that accuracy stability does not necessarily imply identical reasoning or output distributions. 

Stuart-Maxwell tests revealed significant distributional shifts between prompt variants for all models ($\text{p}<0.05$), with the highest performing models showing extremely low p-values ($<10^{-4}$). This suggests that well-performing models, while maintaining stable accuracy, exhibit systematic shifts in their response distribution when prompt syntax changes. Models with lower performance, such as MedFound-7B and ClinicalGPT-base-zh show less coherent changes and potentially more diffuse answers.
From a reproducibility perspective, this pattern highlights that stability in accuracy does not guarantee stability in decision behavior to minimal prompt perturbation.

\begin{table}[!h]
    \centering
    \begin{tabular}{l|cc|cc|cc}
    & \multicolumn{2}{c}{\small{$T=0.3$}} & \multicolumn{2}{c}{\small{$T=0.6$}} & \multicolumn{2}{c}{\small{$T=1.0$}} \\
    {} &  {\checkmark} &  {$\times$}  &  {\checkmark} &  {$\times$}  &  {\checkmark} &  {$\times$}\\
    \hline
    HuatuoGPT-o1-8B &     14 &    11 &      11 &      7  & 10       &  4\\
    {Diabetica-o1}   &      5 &     4 &       7 &      3  & 5        &  3\\
    {Diabetica-7B}    &     13 &    16 &       6 &     13  & 3        & 7 \\
    {Meditron3-8B}    &      2 &     0 &       0 &      0  & 0        & 0 \\
    \end{tabular}
    \caption{Number of ESAP items where the model consistently selected the same option 10 out of 10 runs. The count of correct (\checkmark) and incorrect ($\times$) responses is reported at different temperature regimes.}
    \label{tab:consistency_10_runs}
\end{table}

Across all models, the match rate exceeded Cohen's $\kappa$ by approximately $0.1-0.2$, indicating that a portion of observed agreement reflects shared answer preferences over consistency. Interestingly, Cohen's $\kappa$ and match-rate degraded more strongly when  option letters were removed (Prompt A w vs w/o Letter Token) than when prompt syntax was altered (Prompt A vs Prompt B), suggesting that model consistency depends more on the formatting structure of multiple-choice than on linguistic changes. 

The evaluation of LLMs performance under removal of option letters highlights a strong sensitivity bias toward specific options. For such experiment, Cohen's $\kappa \le 0.4$ for all models. With syntactic variations, only HuatuoGPT-o1-8B and Diabetica-7B show a moderate level of agreement, respectively, with $\kappa=0.55$ and $0.63$.

\subsection{Stability in Non-Deterministic Setting}
HuatuoGPT-o1-8B, Diabetica-o1, Diabetica-7B, and Meditron3-8B emerged as best performing models from Experiment 1 and were used for follow-up analysis assessing the stability of their responses in the non-deterministic setting. 

\begin{figure}[!h]
    \centering
    \includegraphics[width=\linewidth]{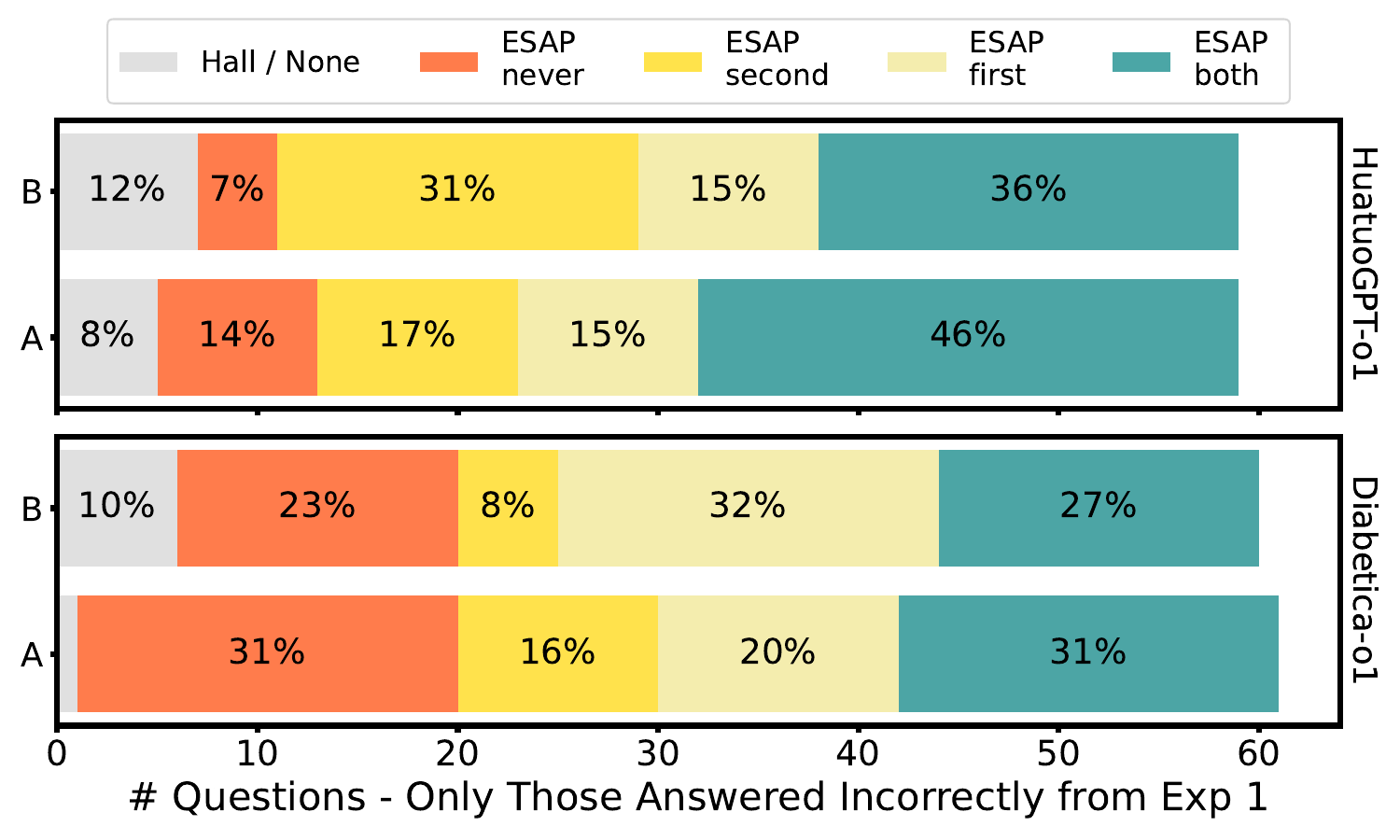}
    \caption{HuatuoGPT-o1-8B and Diabetica-o1 selection of gold-standard explanation against their own reasoning under prompt A and prompt B. 
    Each bar reports the percentage of times each model selected the gold standard explanation, regardless of the presented order (\emph{ESAP both} in green), if only in one of the two positions (\emph{ESAP first} in khaki or \emph{ESAP second} in yellow), in neither (\emph{ESAP never} in red). Cases where a model did not make a valid selection, either by failing to choose between the two explanations or by introducing a new and unrelated option are indicated as \emph{Hallucinate / None}, in gray.}
    \label{fig:Figure2}
\end{figure}

\begin{table}[!ht]
    \centering
    \begin{tabular}{c|c|c|l}
    \hline
         \textbf{Prompt} & \textbf{Case} & \textbf{Score} & \textbf{Main Observations} \\ 
        \hline
          1 & 27 & 3 & Missed Factual Details \\ 
            &    &   & and Key Diagnosis  \\
          \hline
            & 36 & 4 & Acceptable Reasoning \\
          \hline
          & 37 & 4 & Acceptable Reasoning \\
          \hline
          & 57 & 4 & Acceptable Reasoning \\
          &    &   & and Clinical Deviation  \\ 
          \hline
          & 60 & 3 & Partially Correct \\ 
          \hline
          & 71 & 2 & Misclassification of Risk \\
          \hline
          & 75 & 2 & Missed Factual Details\\
          \hline
          & 94 & 3 & Incomplete Reasoning \\
        \hline
        \hline
         2 & 27 & 2 & Wrong Statements \\ 
           &    &   &Missed Factual Details \\
           \hline
        & 34 & 2 & Wrong Statements \\
           \hline
        & 49 & 2 & Misinterpreted Labs \\
           \hline
       & 94 & 2 & Factual Recall Failure\\
    \end{tabular}
    \caption{Endocrinologist evaluation of HuatuoGPT-o1-8B output on questions where model failed to identify the correct reasoning. Responses are evaluated based on clinical reasoning on a Likert scale from 1 to 5.}
    \label{tab:medical_eval}
\end{table}

\begin{table*}[!h]
\centering  
\begin{tabular}{l|ccccc} 
\hline
\textbf{Model} &  \multicolumn{5}{c}{\textbf{Prompt A: CUDA 11.8 vs CUDA 12.8}}  \\
               & $\Delta$ accuracy    & McNemar     &  Stuart-Maxwell & $\kappa$ (95\% CI)  & match-rate (95\% CI) \\
\hline
\hline
HuatuoGPT-o1-8B & $+0.044$   &$0.5$        & $<10^{-4}$     &$0.51[0.38,0.63]$  &$0.60[0.50, 0.70]$ \\ 
Diabetica-o1    & $-0.022$   &$0.8$        & $<10^{-4}$     &$0.53[0.41,0.66]$ &$0.63[0.52,0.72]$ \\
Diabetica-7B    & $+0.011$   &$1.$         & $<10^{-4}$     &$0.68[0.57,0.79]$ &$0.75[0.65,0.83]$ \\
Meditron3-8B    & $+0.011$   &$1.$         & $<10^{-4}$     &$0.31[0.19,0.44]$ &$0.45[0.35,0.55]$ \\
\hline
    \end{tabular}
    \caption{Comparison of LLMs response using same libraries and different CUDA version. $\Delta~\text{ accuracy}=\text{accuracy}_\text{CUDA 12.8}-\text{accuracy}_\text{CUDA 11.8}$. All other metrics are described in Experiment 1 section.}
     \label{tab:cuda_versions}
\end{table*}

In Figure~\ref{fig:Figure1a}, we excluded ESAP items for which the model fails to produce a majority class and report how often a model selected the correct answer versus how often the same model selected its majority class. An ideal model would show all the density concentrated in a single blob at the top-right corner of the plot. In practice, we observe that all tested models deviate from this ideal behavior following different trends. HuatuoGPT-o1-8B shows the highest overall number of consistently correct responses across ten runs for all temperatures (Table~\ref{tab:consistency_10_runs}, $T=0.3$). At maximum consistency, HuatuoGPT-o1-8B is the only model for which we observe an increasing rate of correct responses over incorrect (Table~\ref{tab:consistency_10_runs}, first row). By contrast, Diabetica-o1 and Diabetica-7B exhibit a sharp decline in model consistency for increasing temperature, with Diabetica-7B showing strong consistency for incorrect responses at $T=0.3$. Meditron3-8B becomes more reluctant in providing a final answer at higher temperature values. At $T=1.0$, Meditron3-8B is the only case where the majority vote corresponds to the \emph{Hallucination / None}, for 71 out of 91 outputs, explaining the high density towards the left side of the plot. We report frequencies of the majority class in Supplemental Materials (Tables \ref{tab:majority_T_03},  \ref{tab:majority_T_06}, \ref{tab:majority_T_1}). Table~\ref{tab:consistency_10_runs} further highlights how for HuatuoGPT-o1-8B the temperature increase affects more abruptly the consistency of wrong answers, leaving a relatively higher number of correct answers still highly consistent. 
Overall, these results highlight that consistency, as reflected by output frequency across runs, does not necessarily map to correctness, leading to potential risks of over-relying on output frequency as a proxy for reliability and reproducible behavior. 

We lastly show in Figure~\ref{fig:Figure1b} how the distribution of the number of correct responses (across ten runs) changes for increasing temperatures. While HuatuoGPT-o1-8B accuracy remains stable, the other models show a decline, with Meditron3-8B not providing final answers at $T=1$ and justifying the sharp decline in performance.

\subsection{Discerning Gold-Standard Reasoning}
Our analysis focuses on the cases where HuatuoGPT-o1-8B and Diabetica-o1 responded incorrectly: 61 (prompt A) and 60 (prompt B) for Diabetica-o1, and 59 for both prompts in HuatuoGPT-o1-8B.
Figure~\ref{fig:Figure2}

HuatuoGPT-o1-8B shows higher preference for the gold standard explanation than Diabetica-o1. Under prompt A (and B), the HuatuoGPT-o1-8B selected the gold standard explanation in both positions for 27 (21) cases, selected it in only one position for 19 (26) cases, and it failed to select it in either position for 8 (4) cases. There were 5 (7) cases that could not be classified due to the ambiguous or irrelevant output.
In comparison, Diabetica-o1 selected the gold standard explanation in both positions for 19 (16) cases, in only one position for 22 (24) cases, and never for 19 (14) cases, with 1 (6) non-classifiable responses under prompt 1 (and 2). These differences highlight behavioral reproducibility gaps between the two models: while both show comparable overall accuracy, HuatuoGPT-o1-8B exhibits a greater ability in reasoning alignment with the clinical reference.

A pediatric endocrinology expert reviewed HuatuoGPT-o1-8B outputs where the gold-standard was never selected (12 ESAP items; App. \emph{Integral Evaluations from Pediatric Endocrinologists and HuatuoGPT-o1-8B}. For more detailed clinical comments and comparison with HuatuoGPT-o1-8B output, see Table~\ref{tab:medical_eval}.

The physician noted that HuatuoGPT-o1-8B often overlooked key biochemical findings that would guide a clinician's reasoning and would lead a clinician to immediately consider or rule out specific options. However, only two of the twelve reviewed cases contain factually incorrect statements. This pattern suggests that reasoning might be coherent but incomplete, pointing to systematic rather than random reasoning errors. 

Additional agreement analyses between the two explanations as performed by the LLM are illustrated in Figure~\ref{fig:Figure4a}-\ref{fig:Figure4b} for HuatuoGPT-o1-8B, and Figure~\ref{fig:Figure5a}-\ref{fig:Figure5b} for Diabetica-o1. Those seem to reinforce this view: for HuatuoGPT-o1-8B, the spread of evaluation scores is wider when the model recognizes the ESAP explanation as superior than when it favors its own, suggesting a correlation between self-consistency (selection of ESAP explanation at both positions) and interpretive coherence (evaluation of explanation agreement). We do not observe a similar behavior for Diabetica-o1.

We further refer the reader to Table~\ref{tab:clinician-huatuo-prompt1} and Table~\ref{tab:clinician-huatuo-prompt2} to examine the endocrinologist's evaluation on HuatuoGPT-o1-8B responses. Overall, HuatuoGPT-o1-8B's consider its responses to be more similar to the correct ESAP explanation than the expert physician. This finding is consistent with the model's tendency to overlook key clinical information for certain items.

Overall, the results suggest a high variability for superior explanation selection, dependent on the explanation position within the prompt, but also a coherent, even if flawed, rather than randomic, evaluation of model response against gold-standard for HuatuoGPT-o1-8B.
\subsection{Numerical Instability}
In Table~\ref{tab:cuda_versions}, we report the comparison of LLMs outputs under prompt A, given two different CUDA frameworks. This result shows how across two CUDA versions, all models exhibited statistically significant distributional shifts (Stuart–Maxwell $p<10^{-4}$), despite changes in mean accuracy are still within confidence intervals from Table~\ref{tab:experiment1_accuracy}. Agreement between runs remained moderate at best ($\kappa$ = 0.31–0.68), with match-rates spanning only 0.45–0.75. This divergence, observed in the absence of architectural or dataset changes, suggests that differences in numerical precision and kernel behavior across CUDA builds propagate to measurable inconsistencies in model predictions which lead to completely different clinical outcomes. 

\section{Discussion}
\subsection{Interpretation of Findings}
While the tested small open-source medical LLMs exhibit comparable overall accuracy trends (Experiment 1), they largely differ in the quality, consistency, and behavioral stability of their output under stress-testing. We argue that model assessment should move beyond aggregate performance metrics (such as accuracy) toward finer-grained behavioral diagnostics that capture reasoning stability and behavioral reproducibility, essential for reliable and reproducible science. Despite their medical fine-tuning, the evaluated models show a pronounced bias toward option letters, which becomes particularly problematic in the tested context, suggesting that superficial cues can override semantic reasoning.

As expected, increasing temperature values degraded model performance, reflecting higher creativity and reduced output stability under stochastic sampling~\cite{renze2024effect, li2025exploring}. Although some variation is inherent to LLMs~\cite{atil2024llm}, the degree and pattern of inconsistency varies greatly between models, suggesting that HuatuoGPT-o1-8B could be a better fit to clinical tasks than its counterparts. Our results further suggest that over-reliance on response frequency as a measure of correctness would be misleading. In clinical contexts, such behavior could translate into models persistently recommending inappropriate actions, with the risk of reproducible, yet potentially harmful recommendations.

Even the top-performing model, HuatuoGPT-o1-8B, ehibited self-assessment bias, favoring its own explanations over a provided gold-standard rationale, even when incorrect. Expert review in Experiment 3 showed that some of these outputs were clinically acceptable though not strictly correct or incomplete, whereas in other situations the LLM made critical factual errors. This pattern underscores that current small medical LLMs can exhibit flawed reasoning, emphasizing the need for reproducibility-oriented evaluation frameworks that diagnose not only what LLMs answer, but how they reason.  

Lastly, a further dimension of instability emerged from our numerical reproducibility tests. Even when prompts, model weights, and HuggingFace libraries were held constant, changes in CUDA versions yielded statistically significant distributional shifts in model responses, with moderate rate of agreement. This finding exposes how hardware-level and software-level numerical differences, invisible to end users, can propagate to clinically meaningful output changes. Adding to the multifaceted theme of reproducibility, in exploratory runs we qualitatively observed that identical model checkpoints (CUDA 11.8 setup) produced non-identical outputs when executed through different inference functions within the HuggingFace \texttt{transformers} library, specifically Pipeline and AutoTokenizer+AutoModelForCausalLM. This inconsistency, observed under determistic decoding conditions, highlights how reproducibility can be affected by backend library implementation details within the same software framework. We notice how while the initial tokens for each generation are identical, small syntactic variations appeared later in the sequence, in some cases propagating into semantically distinct completions and different option selection. One example is reported in the Appendix \emph{Reproducibility Issues: pipeline vs AutoTokenizer+AutoModelForCausalLM.generate()}).

\subsection{Limitations}
First, this work focused on prompt variations and temperature changes, which are proven to affect LLM responses even for large models \cite{jeon2025comparative, wang2024prompt, azimi2024accuracy, li2025exploring, renze2024effect}. These parameters were prioritized since they have more widely been discussed and directly linked to reasoning consistency and stochastic robustness. However, other inference hyperparameters were held constant throughout all the experiments, as detailed in the appendix, yet they may also influence model performance. While a more extensive hyperparameter exploration would be valuable, such undertaking remains extremely time-consuming due to human-review, which requires approximately 60 minutes for the evaluation 91 cases.

Additionally, while MCQs provide a scalable framework for benchmarking, our results highlight that LLMs are highly sensitive to prompting. Our analysis of numerical instability was limited in scope and restricted to a subset of cases, nonetheless the results highlight statistical shifts in models behavior. Expanding this line of inquiry will be essential to establish reproducibility standards for clinical AI systems. 

Lastly, we acknowledge that non-expert reviewers may overlook details of medical relevance or subtle hallucinations. Nonetheless, our hybrid strategy mixing non-expert reviews with targeted expert evaluation on responses of difficult interpretation offers a practical compromise between rigor and feasibility. A deeper qualitative assessment of the medical content across all model outputs would be extremely valuable but remains prohibitively resource-intensive. 


\section{Conclusion}
This study highlights that evaluation of small open-source medical LLMs need to move beyond traditional accuracy metrics alone. While traditional metrics remain useful, they obscure deeper sources of variability that affect how reliably LLMs work. By systematically probing prompt sensitivity, response stability, reasoning coherence, and numerical instability we expose multiple dimensions of reproducibility that current evaluation frameworks largely overlook. 

Our findings reveal that even when absolute accuracy appears similar, models differ substantially in their behavioral reproducibility as the ability to produce consistent reasoning and explanation under semantically equivalent conditions. Despite low absolute performance, difficult to contextualize without a human baseline, our analysis reveals how variability in model outputs underscores the need for rigorous and interpretability-focused evaluation frameworks. 

These results also raise concerns about the over-reliance on MCQ-based evaluations which may mask reasoning inconsistency and favor surface-level alignment. Future assessments should incorporate scenario-based assessment or free-text justification grading, allowing for richer and more interpretable insights into model decision processes. OpenAI's recent release of HealthBench \cite{arora2025healthbench} represents a potential area of exploration.
 
Currently, none of the tested small open-source models demonstrated sufficient reliability for clinical decision-making ``as-is''. However, their accessibility underscores the potential for transparent, reproducible, and lightweight architectures when paired with retrieval-augmentation. In future work, we aim to extend these reproducibility-oriented methods toward supporting safe and interpretable model use in low-resource settings.

\section{Acknowledgments}
This work was supported by the National Artificial Intelligence Research Resource Pilot (NAIRR Grant No. 240463) and the Nova Southeastern University President's Research Grant (NSU-PRG 2026). We gratefully acknowledge Paul Gerbino, Daniel Schaible, and Robert Pursell of the Endocrine Society for their interest in this study, and Dr. Veronica Tozzo for her insightful contributions throughout this research development.

\bigskip

\bibliography{aaai2026}
\makeatletter
\@ifundefined{isChecklistMainFile}{
  \newif\ifreproStandalone
  \reproStandalonetrue
}{
  \newif\ifreproStandalone
  \reproStandalonefalse
}
\makeatother

\ifreproStandalone
\documentclass[letterpaper]{article}
\usepackage[submission]{aaai2026}
\setlength{\pdfpagewidth}{8.5in}
\setlength{\pdfpageheight}{11in}
\usepackage{times}
\usepackage{helvet}
\usepackage{courier}
\usepackage{xcolor}
\frenchspacing

\begin{document}
\fi
\setlength{\leftmargini}{20pt}
\makeatletter\def\@listi{\leftmargin\leftmargini \topsep .5em \parsep .5em \itemsep .5em}
\def\@listii{\leftmargin\leftmarginii \labelwidth\leftmarginii \advance\labelwidth-\labelsep \topsep .4em \parsep .4em \itemsep .4em}
\def\@listiii{\leftmargin\leftmarginiii \labelwidth\leftmarginiii \advance\labelwidth-\labelsep \topsep .4em \parsep .4em \itemsep .4em}\makeatother

\setcounter{secnumdepth}{0}
\renewcommand\thesubsection{\arabic{subsection}}
\renewcommand\labelenumi{\thesubsection.\arabic{enumi}}

\newcounter{checksubsection}
\newcounter{checkitem}[checksubsection]

\newcommand{\checksubsection}[1]{%
  \refstepcounter{checksubsection}%
  \paragraph{\arabic{checksubsection}. #1}%
  \setcounter{checkitem}{0}%
}

\newcommand{\checkitem}{%
  \refstepcounter{checkitem}%
  \item[\arabic{checksubsection}.\arabic{checkitem}.]%
}
\newcommand{\question}[2]{\normalcolor\checkitem #1 #2 \color{blue}}
\newcommand{\ifyespoints}[1]{\makebox[0pt][l]{\hspace{-15pt}\normalcolor #1}}

\section*{Reproducibility Checklist}


\checksubsection{General Paper Structure}
\begin{itemize}

\question{Includes a conceptual outline and/or pseudocode description of AI methods introduced}{(yes/partial/no/NA)}
yes

\question{Clearly delineates statements that are opinions, hypothesis, and speculation from objective facts and results}{(yes/no)}
yes

\question{Provides well-marked pedagogical references for less-familiar readers to gain background necessary to replicate the paper}{(yes/no)}
yes

\end{itemize}
\checksubsection{Theoretical Contributions}
\begin{itemize}

\question{Does this paper make theoretical contributions?}{(yes/no)}
no

	\ifyespoints{\vspace{1.2em}If yes, please address the following points:}
        \begin{itemize}
	
	\question{All assumptions and restrictions are stated clearly and formally}{(yes/partial/no)}
	Type your response here

	\question{All novel claims are stated formally (e.g., in theorem statements)}{(yes/partial/no)}
	Type your response here

	\question{Proofs of all novel claims are included}{(yes/partial/no)}
	Type your response here

	\question{Proof sketches or intuitions are given for complex and/or novel results}{(yes/partial/no)}
	Type your response here

	\question{Appropriate citations to theoretical tools used are given}{(yes/partial/no)}
	Type your response here

	\question{All theoretical claims are demonstrated empirically to hold}{(yes/partial/no/NA)}
	Type your response here

	\question{All experimental code used to eliminate or disprove claims is included}{(yes/no/NA)}
	Type your response here
	
	\end{itemize}
\end{itemize}

\checksubsection{Dataset Usage}
\begin{itemize}

\question{Does this paper rely on one or more datasets?}{(yes/no)}
yes

\ifyespoints{If yes, please address the following points:}
\begin{itemize}

	\question{A motivation is given for why the experiments are conducted on the selected datasets}{(yes/partial/no/NA)}
	yes

	\question{All novel datasets introduced in this paper are included in a data appendix}{(yes/partial/no/NA)}
	NA

	\question{All novel datasets introduced in this paper will be made publicly available upon publication of the paper with a license that allows free usage for research purposes}{(yes/partial/no/NA)}
	yes

	\question{All datasets drawn from the existing literature (potentially including authors' own previously published work) are accompanied by appropriate citations}{(yes/no/NA)}
	yes

	\question{All datasets drawn from the existing literature (potentially including authors' own previously published work) are publicly available}{(yes/partial/no/NA)}
	no

	\question{All datasets that are not publicly available are described in detail, with explanation why publicly available alternatives are not scientifically satisficing}{(yes/partial/no/NA)}
	yes

\end{itemize}
\end{itemize}

\checksubsection{Computational Experiments}
\begin{itemize}

\question{Does this paper include computational experiments?}{(yes/no)}
yes

\ifyespoints{If yes, please address the following points:}
\begin{itemize}

	\question{This paper states the number and range of values tried per (hyper-) parameter during development of the paper, along with the criterion used for selecting the final parameter setting}{(yes/partial/no/NA)}
	yes

	\question{Any code required for pre-processing data is included in the appendix}{(yes/partial/no)}
	yes

	\question{All source code required for conducting and analyzing the experiments is included in a code appendix}{(yes/partial/no)}
	yes

	\question{All source code required for conducting and analyzing the experiments will be made publicly available upon publication of the paper with a license that allows free usage for research purposes}{(yes/partial/no)}
	yes
        
	\question{All source code implementing new methods have comments detailing the implementation, with references to the paper where each step comes from}{(yes/partial/no)}
	yes

	\question{If an algorithm depends on randomness, then the method used for setting seeds is described in a way sufficient to allow replication of results}{(yes/partial/no/NA)}
	yes

	\question{This paper specifies the computing infrastructure used for running experiments (hardware and software), including GPU/CPU models; amount of memory; operating system; names and versions of relevant software libraries and frameworks}{(yes/partial/no)}
	yes

	\question{This paper formally describes evaluation metrics used and explains the motivation for choosing these metrics}{(yes/partial/no)}
	yes

	\question{This paper states the number of algorithm runs used to compute each reported result}{(yes/no)}
	yes

	\question{Analysis of experiments goes beyond single-dimensional summaries of performance (e.g., average; median) to include measures of variation, confidence, or other distributional information}{(yes/no)}
	yes

	\question{The significance of any improvement or decrease in performance is judged using appropriate statistical tests (e.g., Wilcoxon signed-rank)}{(yes/partial/no)}
	yes

	\question{This paper lists all final (hyper-)parameters used for each model/algorithm in the paper’s experiments}{(yes/partial/no/NA)}
	yes

\end{itemize}
\end{itemize}
\ifreproStandalone
\end{document}
\fi
\appendix
\onecolumn
\section{Prompting}
In this Section, we include the prompting strategies used in the experiments.
The arguments \texttt{<description>}, \texttt{<question>}, \texttt{<options>} are filled with content from the ESAP 2021-2022 program, with \texttt{<explanation *>} showing two reasoning paths, either the gold-standard explanation extracted from the ESAP or alternatively a model's output from Experiment 1.
\subsection{Prompts in Experiment 1}\label{app:prompts}
\subsubsection{Prompt A}
\emph{‘Given the following clinical case, consider the question, and provide the most likely answer among the provided options and the reasoning process behind your choice. 
\newline
\newline
\texttt{<description>}
\newline
\newline
\texttt{<question>}
\newline
\newline
Options: \texttt{<options>}
\newline
\newline
Please, reduce the output to a paragraph and report the logic steps that led to your answer.'}

\subsubsection{Prompt B}\emph{‘Consider this clinical case, question, and options available.
\newline
\newline
\texttt{<description>}
\newline
\newline
\texttt{<question>}
\newline
\newline
Options: \texttt{<options>}
\newline
\newline
Please do provide your choice among the possible answers. Then provide the reasoning steps behind your choice.'}

\subsection{Prompt in Experiment 3}\label{app:prompt_exp3}
\emph{‘Given the case the following clinical case and multiple choice question, I will provide you two potential explanations. Can you please report the level of agreement from a scale from 0 (null agreement) to 5 (perfect agreement) and select which response makes more clinical sense between the two explanations? 
\newline
\newline
Clinical case: \texttt{<description>}
\newline
\newline
Questions: \texttt{<question>}
\newline
\newline
Options: \texttt{<options>}
\newline
\newline
Explanation 1: \texttt{<explanation 1>}
\newline
\newline         
Explanation 2: \texttt{<explanation 2>}
\newline
\newline         
Please state clearly which is the most clinical correct option between explanation 1 and explanation 2. Then specify the agreement level between explanation 1 and explanation 2.'}


\section{Experimental Setup}
\subsection{Inference and Hyperparameters}\label{app:experimental_setup}

We perform the model inference sequentially, with each model loaded once and used to process the full set of 91 ESAP items in a single session, each as a single batch. By calling the function for output generation separately for each prompt, we assure that the model output is independent from previous prompts. All the experiments are performed using AutoTokenizer+AutoModelForCausalLM+.generate() from the transformers library (version 4.51.3). The experiments ran on a single Nvidia L40S, cuda 11.8. More details on libraries and software versions are available at \\ \url{https://github.com/vanessadamario/SmallMedLLMs-PedEndo/blob/main/requirements.txt}.

\subsubsection{Experiment 1}\label{app:hyper_exp_1}
do\_sample set to False, temperature, top\_p, and top\_k set to None, length\_penalty = 1.0, repetition\_penalty = 1.2, max\_tokens=1500, max\_new\_tokens = 1500, num\_return\_sequences = 1. pad\_token\_id = 128009, eos\_token\_id = 128009, and attention mask are output of the default tokenizer setting, with padding=True and truncation=True. All other arguments are left as default from the AutoTokenizer+AutoModelForCausalLM+.generate().

\subsubsection{Experiment 2}
Same hyperparameters of Experiment 1, with exception of do\_sample = True, temperature=0.6, top\_p=0.9, top\_k=75. 

\subsubsection{Experiment 3}
This shares identical hyperparameters of Experiment 1, with exception of max\_new\_tokens = 2048.

\section{Output Evaluation}\label{app:output_evaluation}

\begin{table}[!h]
    \centering
    \begin{tabular}{llrrrrrrrr}
Model & Prompt &   \emph{A} &   \emph{B} &   \emph{C}  &   \emph{D} &   \emph{E} &  \emph{Hall / None} &  \emph{Multiple}   \\
\hline
\hline
HuatuoGPT-o1-8B             & A                            &  19 &  24 &  23 &  13 &  12 &    0 &     0 \\
                            & B                            &  21 &  19 &  22 &  17 &  12 &    0 &     0 \\
                            & A No Letter                  &  25 &  20 &  19 &  16 &   8 &    1 &     2  \\
\hline
Diabetica-o1                & A                            &  24 &  24 &  20 &  17 &   6 &    0 &     0 \\
                            & B                            &  15 &  23 &  21 &  20 &  12 &    0 &     0 \\
                            & A No Letter                   &  22 &  16 &  22 & 17 &   9 &    3 &     2 \\
\hline
Diabetica-7B                & A                             &  13 &  19 &  28 & 19 &  11 &    1 &     0 \\
                            & B                             &  13 &  21 &  26 & 19 &  12 &    0 &     0 \\
                            & A No Letter                   &  16 &  14 &  18 & 20 &  12 &    5 &      6 \\
\hline
Meditron3-8B                & A                             &  18 &  21 &  18 & 19 &  13 &    2 &     0  \\
                            & B                         &  -  &  -  &  -  &   - &   -  &    - &   - \\
                            & A No Letter               &  17 &  18 &  12 &  19 &  16 &    7 &      2 \\
\hline
MedFound-7B                 & A                         &  13 &   1 &   1 &   1 &   1 &   72 &     2 \\
                            & B                         &  30 &   3 &   1 &   1 &   0 &   56 &     0 \\
                            & A No Opt Letter           &  13 &   3 &   1 &   1 &   2 &   64 &     7 \\
\hline
ClinicalGPT-base-zh         & A                         &  23 &  19 &  17 &  7 &   6 &   16 &     3  \\
                            & B                        &  69 &   2 &   8 &   5 &   2 &    2 &     3 \\
                            & A No Opt Letter       &  46 &   7 &   6 &  7 &   6 &   18 &     1  \\
\hline
\hline
\end{tabular}
    \caption{Experiment 1: Responses distribution for the three prompts tested. MedFound-7B and ClinicalGPT-base-zh show the highest degree of hallucination. Meditron3-8B does not output a response under prompt B.}
    \label{tab:my_label}
\end{table}

\begin{table}[!h]
    \centering
\begin{tabular}{llrrrrrrrr}
Model &  Temperature &  \emph{A} &    \emph{B} &    \emph{C} &   \emph{D} &    \emph{E} &  \emph{Hall / None} &  \emph{Multiple}  \\
\hline
\hline
HuatuoGPT-o1-8B & 0.3     &  172 &  236 &  206 &    168 &  120 &            5 &      3 \\
                & 0.6     &  171 &  231 &  201 &    171 &  124 &           12 &      0 \\
                & 1.0     &  171 &  234 &  196 &    170 &  120 &           17 &      2\\
\hline
Diabetica-o1    & 0.3     &  203 &  210 &  219 &    172 &  106 &            0 &      0 \\
                & 0.6     &  204 &  210 &  215 &    173 &   97 &           11 &      0 \\
                & 1.0     &  181 &  213 &  224 &    175 &   98 &           12 &      7 \\
\hline
Diabetica-7B    & 0.3     &  146 &  170 &  276 &    194 &  113 &            8 &    3 \\
                & 0.6     &  170 &  166 &  252 &    198 &  110 &           12 &      2 \\
                & 1.0       &  158 &  169 &  237 &  162 &  123 &           53 &    8 \\
\hline
Meditron3-8B    & 0.3    &  127 &  195 &  202 &     195 &  164 &           26 &      1 \\
                & 0.6   &  149 &  202 &  182 &      182 &  156 &           39 &      0  \\
                & 1.0   &   82 &  111 &  109 &      70 &   56 &          476 &      6  \\
\hline\hline
\end{tabular}
    \caption{Experiment 2: Responses distribution across ten runs of prompt A.}
    \label{tab:readings_exp_2}
\end{table}

\begin{table}[!h]
    \centering
    \begin{tabular}{lrrrrrrr}
    {} &   \emph{A} &   \emph{B} &   \emph{C} &   \emph{D} &   \emph{E} &  No Majority \\
    \hline
    \hline
    HuatuoGPT-o1-8B &  17 &  24 &  17 &  16 &  10 &    7 \\
    \hline
    Diabetica-o1    &  18 &  20 &  22 &  13 &   8 &   10 \\
    \hline
    Diabetica-7B    &  13 &  18 &  29 &  17 &   9 &    5 \\
    \hline
    Meditron3-8B    &  11 &  19 &  20 &  17 &  13 &   11 \\
    \hline
    \hline
    \end{tabular}
    \caption{Frequency of the majority class for $T=0.3$.}
    \label{tab:majority_T_03}
\end{table}

\begin{table}[!h]
    \centering
    \begin{tabular}{lrrrrrrr}
    {} &   \emph{A} &   \emph{B} &   \emph{C} &   \emph{D} &   \emph{E}  &  No Majority \\
    \hline
\hline
    HuatuoGPT-o1-8B &  16 &  22 &  21 &  17 &  12 &    3 \\
    \hline
    Diabetica-o1    &  22 &  18 &  23 &  16 &   7 &    5 \\
    \hline
    Diabetica-7B    &  16 &  15 &  27 &  19 &   9 &    5 \\
    \hline
    Meditron3-8B    &   9 &  23 &  16 &  17 &  11 &   15 \\
    \hline
    \end{tabular}
    \caption{Frequency of majority class for $T=0.6$.}
    \label{tab:majority_T_06}
\end{table}

\begin{table}[!h]
    \centering
    \begin{tabular}{lrrrrrrr}
    {} &   \emph{A} &   \emph{B} &   \emph{C} &   \emph{D} &   \emph{E} &  \emph{Hall / None} &  No Majority \\
    \hline
\hline
    HuatuoGPT-o1-8B &  17 &  23 &  20 &  15 &  11 &     0 &    5 \\
    \hline
    Diabetica-o1    &  13 &  19 &  25 &  16 &   8 &     0 &   10 \\
    \hline
    Diabetica-7B    &  14 &  19 &  26 &  15 &  11 &     0 &    6 \\
    \hline
    Meditron3-8B    &   3 &   4 &   0 &   2 &   2 &    71 &    9 \\
    \hline
\hline
    \end{tabular}
    \caption{Frequency of majority class for $T=1.0$. Meditron3-8B avoid option selection for most cases (71 out of 91).}
    \label{tab:majority_T_1}
\end{table}


\section{Visualization of Model Consistency Across Temperatures}

\begin{figure}[!h]
    \centering
        \includegraphics[width=\linewidth]{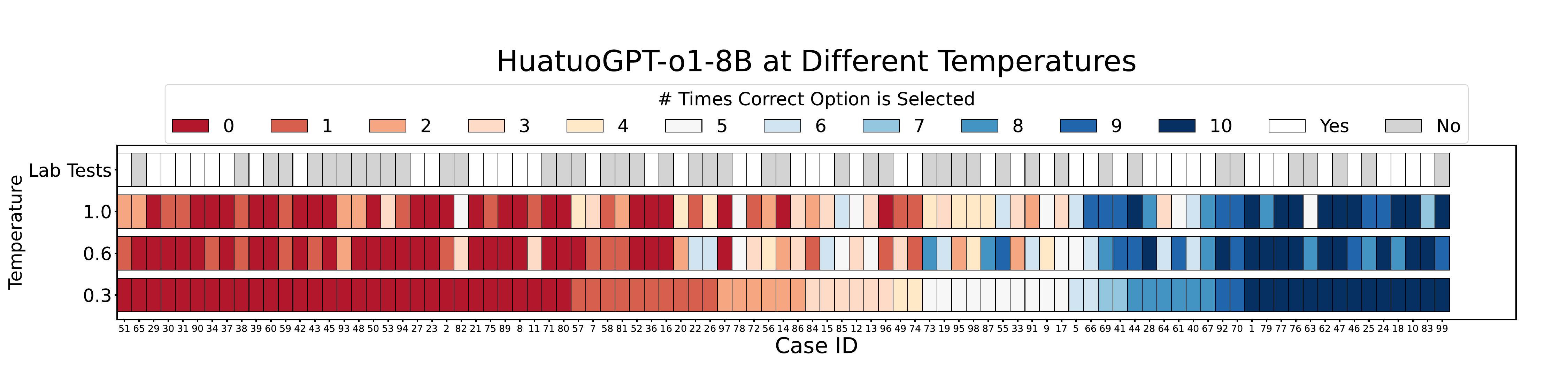}
        \caption{Graphical representation of HuatuoGPT-o1-8B consistency and correctness at different temperature values.}
        \label{fig:Figure3a}
    \end{figure}

\begin{figure}[!h]
        \includegraphics[width=\linewidth]{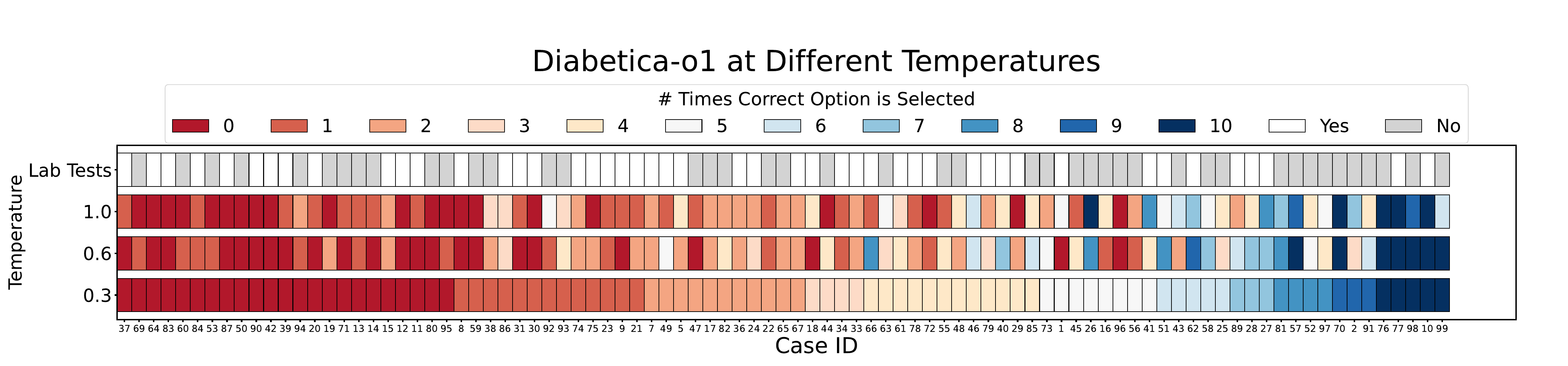}
        \caption{Graphical representation of Diabetica-o1 consistency and correctness at different temperature values.}
        \label{fig:Figure3b}
    \end{figure}

\begin{figure}[!h]
        \includegraphics[width=\linewidth]{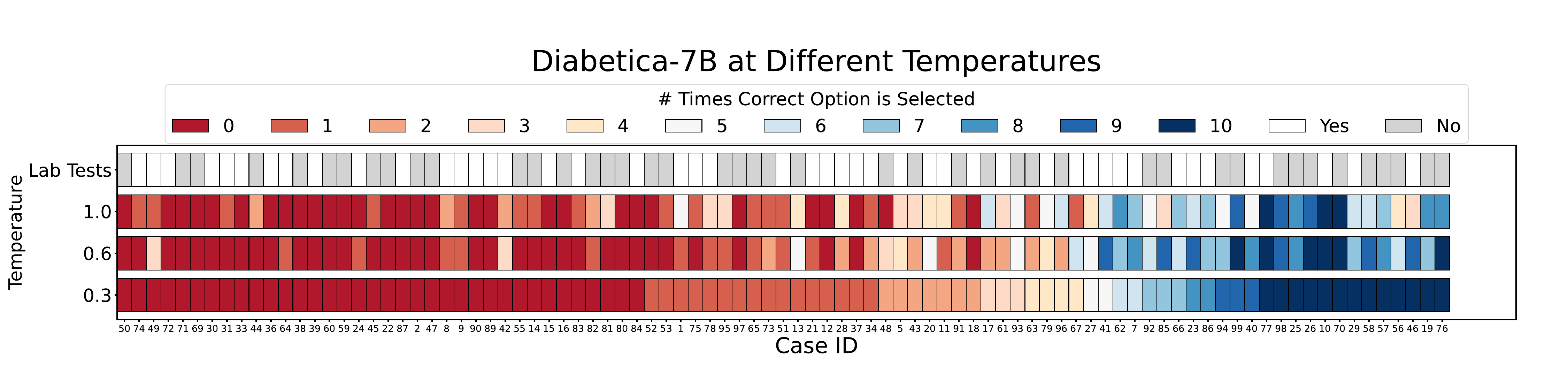}
        \caption{Graphical representation of Diabetica-7B consistency and correctness at different temperature values.}
        \label{fig:Figure3c}
    \end{figure}

\begin{figure}[!h]
        \includegraphics[width=\linewidth]{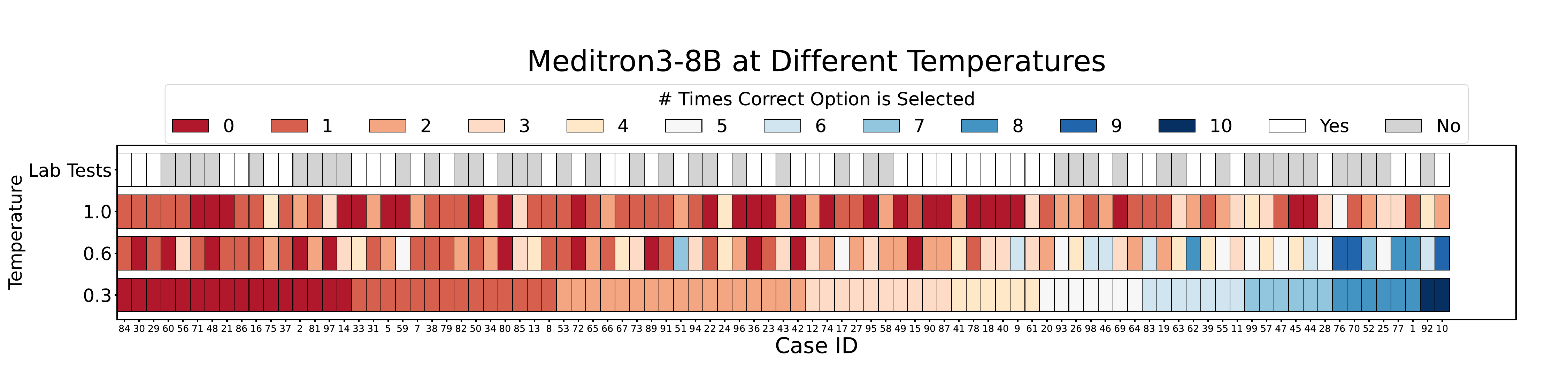}
        \caption{Graphical representation of Meditron3-8B consistency and correctness at different temperature values.}
        \label{fig:Figure3d}
    \end{figure}

Figures~\ref{fig:Figure3a}-\ref{fig:Figure3d} show how model correctness changes across different temperatures, for the three $T$ values tested. ESAP items are ordered based on each model performance at the lowest temperature tested ($T=0.3$). HuatuoGPT-o1-8B exhibits the highest coherence. Diabetica-7B and Diabetica-o1 represent the intermediate case. Interestingly, Diabetica-7B shows much higher consistency and correctness for $T=0.3$ but its performance degrades quite abruptly for increasing temperature. Meditron3-8B, as discussed in the main body of the paper, become more resistant in providing a response for $T=1.0$, which explains the lack of higher numerical values. 

\section{Further Results for Experiment 3: Discerning Gold-Standard Reasoning}
\subsection{Agreement between Gold Standard and Models Outputs from a Model's Perspective}\label{app:exp3}
We extracted agreement scores from the model outputs on a 0–5 scale. For each case, we calculated the average agreement by combining the scores of two runs: one in which the gold standard explanation was presented first and the model’s reasoning second, and another with the order reversed. Since agreement is less informative when the model already selects the correct answer, our analysis focuses on cases where the model’s selected option disagreed with the gold standard choice. The goal of the following evaluation is to observe whether the level of agreement reflects the discrimination capabilities of the model to select the gold standard explanation against the model's own reasoning. We excluded the clinical cases where the model's reasoning selects a different answer from the one in explanation 1 and explanation 2.

\begin{figure}[h]
  \centering
    \centering
    \includegraphics[width=0.33\textwidth]{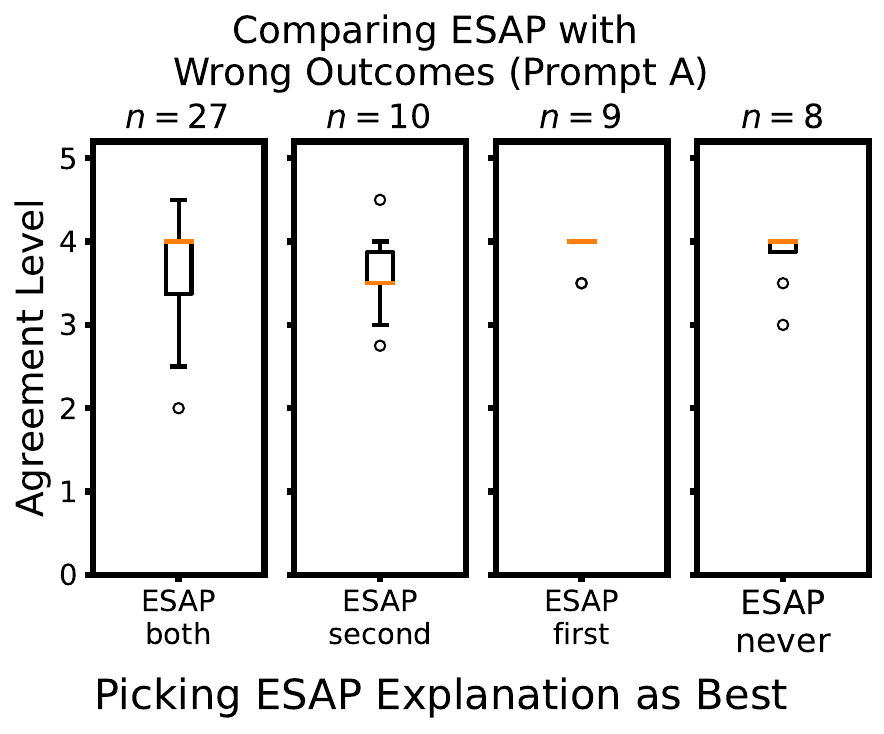}
    \caption{Distribution of agreement scores for \textbf{HuatuoGPT-o1-8B}: Prompt A}
    \label{fig:Figure4a}
\end{figure}

\begin{figure}[h] 
    \centering
    \includegraphics[width=0.33\textwidth]{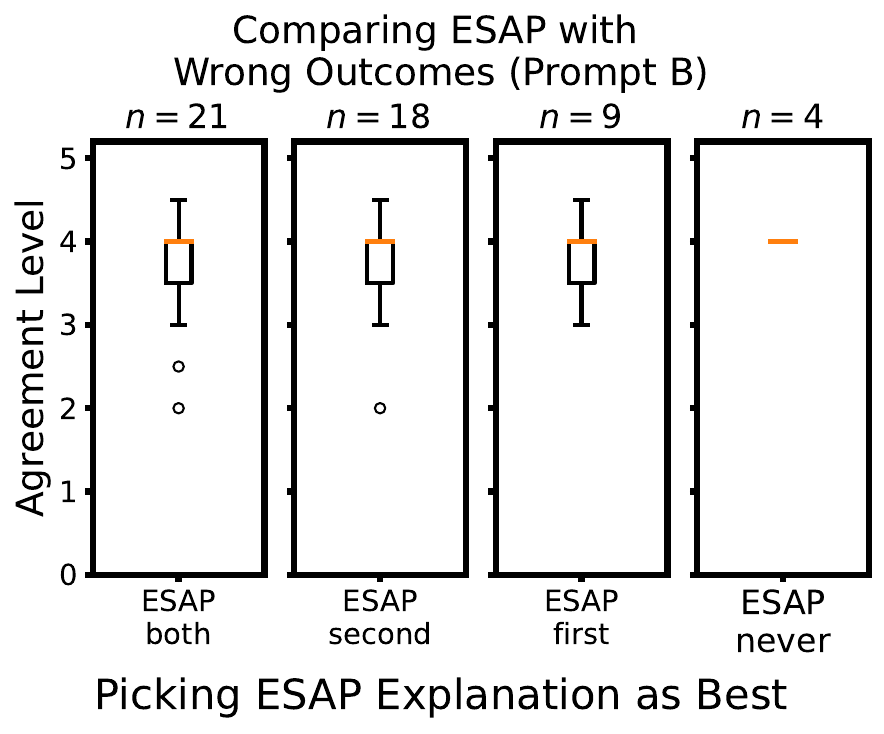}
    \caption{Distribution of agreement scores for \textbf{HuatuoGPT-o1-8B}: Prompt B}
    \label{fig:Figure4b}
\end{figure}

\begin{figure}[h]
  \centering
    \centering
    \includegraphics[width=0.33\textwidth]{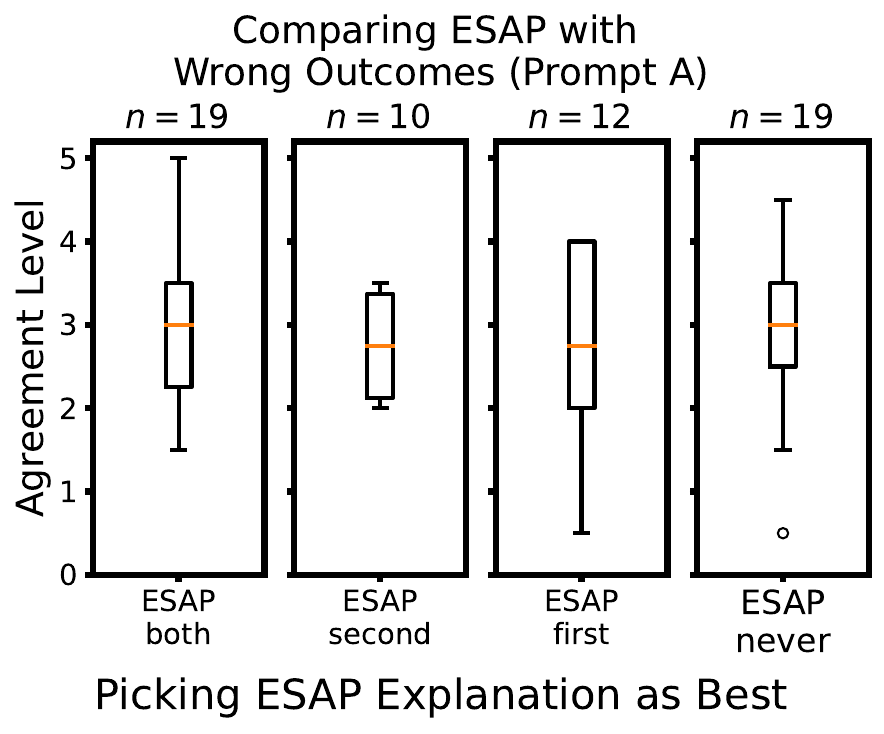}
    \caption{Distribution of agreement scores for \textbf{Diabetica-o1}: Prompt A}
    \label{fig:Figure5a}
\end{figure}

\begin{figure}[h] 
    \centering
    \includegraphics[width=0.33\textwidth]{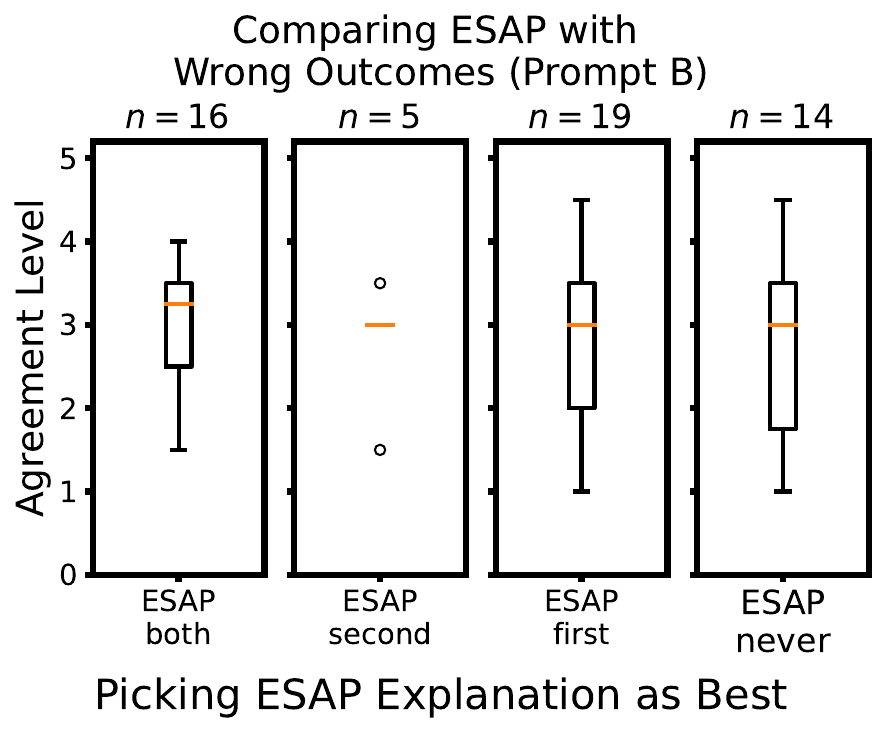}
    \caption{Distribution of agreement scores for \textbf{Diabetica-o1}: Prompt B}
   \label{fig:Figure5b}
\end{figure}

Figures \ref{fig:Figure4a}-\ref{fig:Figure4b} and Figures \ref{fig:Figure5a}-\ref{fig:Figure5b} show respectively the average agreement score for HuatuoGPT-o1-8B and Diabetica-o1 for prompt A and prompt B.
  
The analysis can only be speculative given the small sample size. 
While Diabetica-o1 does not show any specific trend, the box-plot distributions for HuatuoGPT-o1-8B suggest something interesting. In Figures~\ref{fig:Figure4a}-\ref{fig:Figure4b}, agreement scores are more widely distributed for cases where the model selected the gold standard explanation in both runs (Both), indicating greater variability of the agreement score and the potential capability of  discerning clinical differences between the gold standard and the misleading response of the model itself. Interestingly, when the model favored its own reasoning over the gold standard (Never) ($n=8$ in prompt 1 and $n=4$ in prompt 2), the box plots are generally more skewed towards higher agreement values, potentially reflecting a lack of sensitivity to the subtle distinctions in the gold standard explanation versus the model reasoning.
The values in the right plot (Never) of Figure~\ref{fig:Figure4a} and Figure~\ref{fig:Figure4b} can be qualitatively compared to the scores in Table~\ref{tab:medical_eval}, where the evaluation based on clinical reasoning is on a 1-5 scale. HuatuoGPT-o1-8B shows higher levels of agreement than the ones assigned by the pediatric endocrinology expert. Evaluation of experiment 1 outcomes is given by the pediatric endocrinologist and HuatuoGPT-o1-8B (from Experiment 3).

\subsection{Integral Evaluations from Pediatric Endocrinologist and HuatuoGPT-o1-8B}\label{app:clinical_further_eval}

Table~\ref{tab:clinician-huatuo-prompt1} and Table~\ref{tab:clinician-huatuo-prompt2} show the clinician's evaluation of HuatuoGPT-o1-8B responses of Experiment 1 and their evaluation in Likert scale (1-5) based on the clinical reasoning scale in CLEVER. The right side of the tables shows the evaluation of HuatuoGPT-o1-8B's previous explanation against the gold standard. In general, we notice that HuatuoGPT-o1-8B tends to be more forgiving, losing some important details related to biochemical and genetic mechanisms which are key for the diagnosis. 

\begin{table}[!h]
    \centering
    \begin{tabular}{c||p{4cm}|c||p{4.cm}|p{4.cm}|c}
    \hline
        Case & \textbf{Pediatric Endocrinology Expert} &(1-5) & \textbf{HuatuoGPT-o1-8B} \footnotesize Explanation 1: Gold-Standard; Explanation 2: Model Reasoning & \footnotesize Explanation 1: Model Reasoning; Explanation 2: Gold-Standard& (0-5)\\ 
        \hline
         27 & \footnotesize Reasoning is generally correct, but did not get to the biochemical diagnosis of 5-$\alpha$ reductase deficiency. & 3 &\footnotesize [..]. explanation 1 focusing on genetic considerations and explanation 2 emphasizing physiological mechanisms. & \footnotesize [..] their approaches differ slightly in interpreting the significance of each detail. & 4 \\
         36 & \footnotesize Answer is actually not totally unreasonable and honestly I would consider doing this as well. & 4 & \footnotesize [..] While Explanation 1 prioritizes safety, Explanation 2 considers ongoing treatment efficacy. & \footnotesize [..] While explanation 1 focuses primarily on adjusting the medication dosage, explanation 2 provides additional context and potential treatment alternatives.  & 4 \\
         37 &  \footnotesize This question is a bit tricky  as it asks about order. I may do an ACTH stimulation test as well. [..] The model actually considered the correct answer initially, so the thinking is generally correct here. & 4 & \footnotesize [..] while Explaining 2 offers practical insights quickly, Explaining 1 covers broader diagnostic ground. They complement each other nicely. & \footnotesize [..]  There is partial overlap in recommendations regarding the necessity of initial adrenal function testing, leading to a moderate level of agreement. & 4  \\
         57 & \footnotesize Again things were generally correct here. Some medical centers and if the patient was in the hospital, honestly we would probably do the stimulation test. & 4 & \footnotesize [..] Both perspectives recognize the critical nature of assessing adrenal status post-long term steroid exposure but differ slightly in methodology preference. & \footnotesize [..] Explanation 1 better suits scenarios demanding prompt resolution, whereas Explanation 2 fits better in non-emergency settings where gradual tapering is feasible. & 4 \\ 
         60 & \footnotesize Hormone profile should be that of someone in the fed state, but the hypothalamus cannot see those signals. It got some of it correct, but not all.  & 3 & \footnotesize [..] They differ in specific detail [..]. Although both capture key aspects of the condition, they present distinct interpretations of the same underlying problem.& \footnotesize [..] While insightful, this approach (\texttt{ESAP gold-standard response}) does not directly address the immediate concerns presented in the initial scenario. & 3.5 \\ 
         71 & \footnotesize Model did not assign that despite the large thyroid nodule, it is actually a low risk cancer and therefore the TSH does not need to be as aggressively suppressed. & 2 & \footnotesize Explanation 1 [..] provides context essential for understanding the nuances involved in risk categorization. Meanwhile, explanation 2 distills this complexity into actionable steps, emphasizing the critical balance needed in real-world scenarios.& \footnotesize [..] **Explanation 1 is slightly more directly applicable** due to its focus on general principles rather than strict categorization rules. However, **both explanations share significant relevance** [..]. & 4  \\
         75 & \footnotesize The model did not get to the key part of this question which is the actual biochemical profile that was described. I would not be able to get this correct if I could not have access to the internet as this is really a question about facts. & 2 & \footnotesize [..] Explanation 1 provides a detailed background on each gene. [..] However, it feels a bit dry and lacks the real-world application aspect. Meanwhile, explanation 2 gets straight to the point. & \footnotesize [..] Explanation 1 zeroes in on the specifics of the case, whereas Explanation 2 takes a broader view of genetic considerations. & 3 \\
         94 & \footnotesize There were scattered truths here. Again, this is a tough question at the end of the day that I would struggle with. I just don't think that it got to the crux of the question, what is the most common pathogenic genetic change that we see in kids. & 3 & \footnotesize [..] Initially, I leaned towards \texttt{ESAP gold-standard response} due to its general popularity, but upon further consideration, the stronger link to pediatric cases made \texttt{model's response} the better choice.& \footnotesize [..] Both explanations recognize the importance of (\texttt{ESAP gold-standard response}) and (\texttt{model's response}) in papillary thyroid cancer, but differ in their prioritization based on the patient's age and disease characteristics. & 4 \\
         \hline
    \end{tabular}
    \caption{HuatuoGPT-o1-8B reasoning for incorrect answers to prompt 1 where model consistently failed to identify gold-standard reasoning (both). On the left, pediatric endocrinology expert evaluation, on the right, Huatuo-o1-8B evaluation.}
\label{tab:clinician-huatuo-prompt1}
\end{table}

\begin{table}[!h]
    \centering
    \begin{tabular}{c||p{4cm}|c||p{4.cm}|p{4.cm}|c}
    \hline
        Case & \textbf{Pediatric Endocrinology Expert} &(1-5) & \textbf{HuatuoGPT-o1-8B} \footnotesize Explanation 1: Gold-Standard; Explanation 2: Model Reasoning & \footnotesize Explanation 1: Model Reasoning; Explanation 2: Gold-Standard& (0-5)\\ 
        \hline
         27 & \footnotesize There is worse reasoning here and wrong statements about hormone levels that are in the question stem. Again, it does not get to the 10:1 ratio that is needed for 5-$\alpha$ RD. & 2  &\footnotesize [..] Both explanations touch on similar themes regarding androgen function, but they differ slightly in their interpretations of the exact mechanisms involved [..] they diverge somewhat in interpreting the precise mechanisms. & \footnotesize [..] I’d say explanation 1 captures the essence of the biochemical puzzle better, focusing precisely on the enzyme-related aspect. Yet, explanation 2 offers a comprehensive view incorporating familial patterns and clinical manifestations. &  4 \\
         34 & \footnotesize Honestly, strange question. Insulinoma is extremely rare at this age. Diazoxide challenge really is not useful that it picked. There are physiology statements that are not exactly true. It did have the correct answer interestingly and almost chose it. & 2  &\footnotesize [..] However, slight variations exist in the emphasis placed on each detail, reflecting nuanced differences in interpretation. Overall, the conclusions drawn share significant overlap. & \footnotesize  Although both paths explore the same core issue, they differ in emphasis and scope. Explanation 1 prioritizes functional diagnostics, while Explanation 2 leans towards structural assessments and genetic underpinnings. & 4 \\
         49 & \footnotesize There was some misinterpretation of the lab results here. There is no evidence of hypothyroidism at this point. It sounds like the model had concerns about the large drop in T4. There is no evidence of over-treatment. & 2  &\footnotesize [..] I'd say Explanation 2 is the more clinically sound choice. Its emphasis on timely intervention to maintain optimal thyroid function resonates with real-world medical decision-making. & \footnotesize [..] Both explanations recognize the significance of the T4 decline and its implications for the patient's treatment trajectory. They share similar insights regarding the necessity of avoiding over-treatment and ensuring safety, although they differ slightly in their prioritization of specific diagnostic steps. &  4 \\
         94 & \footnotesize There was some misinterpretation of the lab results here. There is no evidence of hypothyroidism at this point. It sounds like the model had concerns about the large drop in T4. There is no evidence of over-treatment.  & 2  &\footnotesize [..] the most clinically sensible conclusion regarding the genetic alteration causing this papillary thyroid carcinoma in the 16-year-old patient is \texttt{model's response}, due to its higher frequency in pediatric cases compared to \texttt{ESAP gold-standard response}. & \footnotesize [..] it becomes clear that \texttt{ESAP gold-standard response} remains the top contender for the underlying genetic issue in this adolescent case. Despite initial thoughts favoring \texttt{model's response}, the data points strongly toward \texttt{ESAP gold-standard response}as the most clinically relevant factor. [..] the most clinically sensible explanation is **Explanation 1**. & 4 \\
         \hline
    \end{tabular}
    \caption{HuatuoGPT-o1-8B reasoning for incorrect answers to prompt 2 where model consistently failed to identify gold-standard reasoning (both). On the left, pediatric endocrinology expert evaluation, on the right, Huatuo-o1-8B evaluation.}
    \label{tab:clinician-huatuo-prompt2}
\end{table}

\section{Reproducibility Issues: pipeline vs AutoTokenizer + AutoModelForCausalLM.generate()}\label{app:pipeline_vs_tokenizer}
In this study, all small open-source LLMs were run using the HuggingFace \colorbox{gray!15}{\texttt{transformers}} library. Text generation was performed using two approaches: the high-level \colorbox{gray!15}{\texttt{pipeline()}} interface and a lower-level setup involving \colorbox{gray!15}{\texttt{AutoTokenizer}}, \colorbox{gray!15}{\texttt{AutoModelForCausalLM}}, and manual use of \colorbox{gray!15}{\texttt{.generate()}}. Both were configured with \texttt{do\_sample=False} and a \texttt{temperature=0}, which should theoretically result in deterministic result. Between the two workflows, only the pad\_token\_id and eos\_token\_id parameters are different, but our understanding is that, since we avoid the use of batches and call the function for output generation separately for each prompt, those tokens should not affect the model generation. 

Although in some cases the two methods produced identical responses, we observed instances where the generated outputs diverged after the initial few tokens, despite identical prompts and generation settings. In the manual \colorbox{gray!15}{\texttt{.generate()}} setup, we explicitly set \texttt{pad\_token\_id} and \texttt{eos\_token\_id} to the model’s end-of-sequence token. However, these parameters are not exposed or controlled in the same way through \colorbox{gray!15}{\texttt{pipeline()}}, which can lead to implicit differences in how padding or attention masks are handled during inference.

This divergence highlights how small numerical variations might cause significant differences in auto-regressive generation, even under deterministic decoding conditions. This invites further considerations when evaluating model consistency and results reproducibility.




\noindent
\begin{minipage}[t]{0.48\textwidth}
\fbox{%
  \begin{minipage}[t]{\textwidth}
  \small
  \textbf{{AutoTokenizer + AutoModelForCausalLM}:} \\

\footnotesize \#\# Thinking

\vspace{0.2cm}
\footnotesize Alright, let's think about this little guy who just came in with some pretty interesting symptoms. So first off, we've got him having trouble with his eyes - specifically, there's something going on with his vision where one eye turns inward. That happened right after he took quite a tumble when he fell while trying to stand up by himself. Now, they did an MRI and found that both of his optic nerves look swollen, kind of like they're twisted around too much, and even the back part of his eyeballs seem a bit flat. This all sounds like it could be related to increased pressure inside his skull, maybe because of fluid buildup or swelling somewhere. 

\vspace{0.2cm}
\footnotesize Now, moving onto what else is happening with him... Oh, I see! \textit{They mentioned he has an unusual head shape. It seems like his skull isn't developing as expected, especially across the top where you'd expect two separate parts but instead, everything fused together into more of a long oval shape. And then there's how tall he is – way shorter than other kids his age, definitely below the fifth percentile. Plus, he's got these issues with his legs, like bowing and a funny walk. Hmm, those sound familiar...}

\vspace{0.2cm}
\textit{[..]}

\vspace{0.2cm}
\footnotesize \#\# Final Response

\vspace{0.2cm}
\textit{\footnotesize The clinical findings described in the scenario point towards a skeletal disorder characterized by disproportionate short stature, limb deformities, and craniofacial anomalies. Given the presence of short stature ($\mathit{<5^{\text{th}}}$ percentile), mild genu varum (bow-leggedness) deformity, and a waddling gait, along with the distinct head shape indicating complete fusion of the sagittal suture and dolichocephaly, \textbf{the most likely underlying skeletal disorder is **C. Hypochondroplasia**}. This diagnosis aligns with the characteristic features observed in the patient, particularly the absence of dental abnormalities and the nature of the limb deformities. Other potential diagnoses, such as vitamin D deficiency or genetic syndromes, do not fully account for the specific constellation of symptoms presented. Therefore, hypochondroplasia emerges as the most plausible explanation for the child's condition.}

\end{minipage}
}
\end{minipage}
\hfill
\begin{minipage}[t]{0.48\textwidth}
\fbox{%
  \begin{minipage}[t]{\textwidth}
  \small
  \textbf{{Pipeline}}: \\
  
    \footnotesize \#\# Thinking

\vspace{0.2cm}
\footnotesize Alright, let's think about this little guy who just came in with some pretty interesting symptoms. So first off, we've got him having trouble with his eyes - specifically, there's something going on with his vision where one eye turns inward. That happened right after he took quite a tumble when he fell while trying to stand up by himself. Now, they did an MRI and found that both of his optic nerves look swollen, kind of like they're twisted around too much, and even the back part of his eyeballs seem a bit flat. This all sounds like it could be related to increased pressure inside his skull, maybe because of fluid buildup or swelling somewhere. 

\vspace{0.2cm}
\footnotesize Now, moving onto what else is happening with him... Oh, I see! \textit{There’s more than just the eyes. They mentioned he has a funny-shaped head – not round but sort of long and narrow, almost like a football. And then there's how tall he isn't growing compared to other kids his age. Plus, he's got these bowed legs and walks with a weird, wobbly gait. It seems like everything points towards some bone issues here. Let me think...}

\vspace{0.2cm}
\textit{[..]}



\vspace{0.2cm}
\footnotesize \#\# Final Response

\vspace{0.2cm}
\textit{\footnotesize The clinical presentation described suggests several key features: bilateral eye issues post-fall, unusual cranial shape, short stature, leg bowing, and a distinctive gait. Given these details, \textbf{the condition that aligns most closely with these findings is **D. Severe vitamin D deficiency**}. This conclusion is supported by the presence of rachitic changes seen on x-ray, indicating poor bone mineralization typically caused by insufficient vitamin D intake during critical growth periods. While other conditions such as hypochondroplasia may share similar skeletal manifestations, the specific context of inadequate vitamin D supplementation early in life strongly supports severe vitamin D deficiency as the primary concern. Other options listed do not fully account for the comprehensive symptom set presented. Therefore, option D is the most fitting diagnosis based on the information provided.}
  \end{minipage}
}
\end{minipage}

\end{document}